\documentclass[11pt,a4paper]{article}
\usepackage[hyperref]{acl2020}
\usepackage{times}
\usepackage{latexsym}
\usepackage{makecell}
\usepackage{booktabs}
\usepackage{xcolor}
\usepackage{tabu}
\usepackage{array,multirow}
\usepackage{graphicx}

\usepackage{enumitem}
\usepackage{microtype}
\usepackage{CJKutf8} 

\aclfinalcopy 

\title{NEJM-enzh: A Parallel Corpus for English-Chinese Translation in the Biomedical Domain}

\author{Boxiang Liu \\
  Baidu Research \\
  \texttt{jollier.liu@gmail.com} \\\And
  Liang Huang \\
  Baidu Research \& Oregon State University \\
  \texttt{liang.huang.sh@gmail.com} \\}

\date{}

\begin{document}
\begin{CJK}{UTF8}{gbsn}
\maketitle
\begin{abstract}
Machine translation requires large amounts of parallel text. While such datasets are abundant in domains such as newswire, they are less accessible in the biomedical domain. Chinese and English are two of the most widely spoken languages, yet to our knowledge a parallel corpus in the biomedical domain does not exist for this language pair. In this study, we develop an effective pipeline to acquire and process an English-Chinese parallel corpus, consisting of about 100,000 sentence pairs and 3,000,000 tokens on each side, from the New England Journal of Medicine (NEJM). We show that training on out-of-domain data and fine-tuning with as few as 4,000 NEJM sentence pairs improve translation quality by 25.3 (13.4) BLEU for en$\to$zh (zh$\to$en) directions. Translation quality continues to improve at a slower pace on larger in-domain datasets, with an increase of 33.0 (24.3) BLEU for en$\to$zh (zh$\to$en) directions on the full dataset.

\end{abstract}

\section{Introduction}

Recent advances in machine translation have demonstrated translation quality arguably on par with professional human translators in select domains \citep{hassan2018achieving}. Supervised training of machine translation models usually benefit from large amounts of parallel corpora, and such effect is the most evident for neural machine translation models. However, the collection and alignment of parallel corpora requires significant time and labor and such datasets are not available for all domains or langauge pairs.

Machine translation in the biomedical domain is characterized by a long tail of medical terminology. For example, the Unified Medical Language System developed by the National Institute of Health contains over 2 millions names for over 900,000 concepts \citep{bodenreider2004unified}, much larger than the set of common English words used for daily communication. Because neural machine translation models are vulnerable to rare words \citep{luong2014addressing,sennrich2015neural}, their ability to generalize to the biomedical domain, when trained otherwise, is low.

\begin{figure}[t!]
\includegraphics[width=\columnwidth]{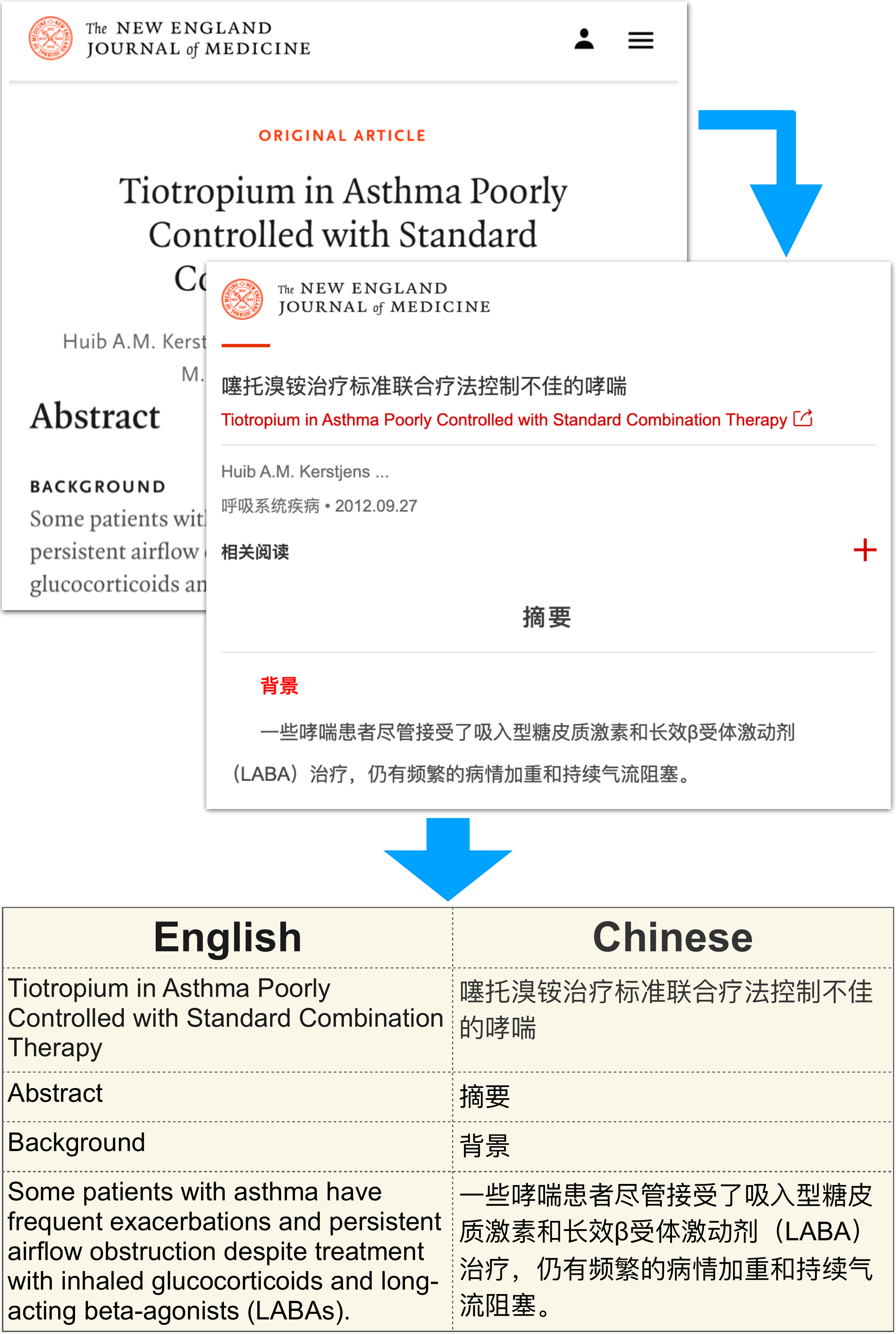}
\caption{An overview the NEJM-enzh corpus construction. The inputs are NEJM webpages and the output is a Chinese/English parallel corpus.}
\label{fig:overview}
\end{figure}

While the need for biomedical parallel corpora is evident, they are not available for all language pairs. At the time of writing, the latest shared task on biomedical translation from WMT19 provides test data on five language pairs (en$\leftrightarrow$zh, en$\leftrightarrow$fr, en$\leftrightarrow$de, en$\leftrightarrow$pt, en$\leftrightarrow$es). Among these five pairs, \textbf{\textit{in-domain}} training data was provided for four pairs but not for en$\leftrightarrow$zh \citep{bawden2019findings}. Although a large amount of matching biomedical English and Chinese texts are freely available on the internet (e.g. PubMed), there has been a lack of coordinated effort to curate such resources into an accessible format for machine translation.

The New England Journal of Medicine (NEJM) provides free and official Chinese translations of its publications dating back to 2011 (\url{http://nejmqianyan.cn/}). The website repository currently hosts nearly 2000 articles, with new articles added weekly. These articles include original research articles, clinical case reports, review articles, commentaries, Journal Watch (viz.~article highlights), and etc. The articles are translated by professional translators and proofread by members of the NEJM editorial team. For research articles, translations on statistical analysis are proofread by statisticians who are native Chinese speakers.

In this study, we present a English-Chinese parallel corpus in the biomedical domain constructed from NEJM (\autoref{fig:overview}) . We provide sentence-aligned bitext for 1,973 articles, totaling 97,441 sentences, and 3,028,434/2,916,772 English/Chinese tokens. In addition, we show that training a baseline model with WMT18 newswire data \citep{barrault2019findings} and fine-tuning the model with the NEJM-enzh dataset will significantly improve translation quality over the baseline model.

Our contributions are the following:
\begin{itemize}[noitemsep,topsep=0pt]
\setlength\itemsep{1em}
\item We present the first English-Chinese parallel corpus in the biomedical domain (\autoref{sec:data availability}). This is the main contribution of our paper. 

\item We compare several software packages for sentence boundary detection and alignment. Our experiments show that widely used open-source packages may be suboptimal in the biomedical domain (\autoref{sec:corpus construction}).
\item We show that fine-tuning on as few as 4,000 sentence pairs from NEJM-enzh can improve translation quality by 25.3 (13.4) BLEU for en$\to$zh (zh$\to$en). Translation quality continues to improve at a slower pace on larger datasets, finishing at an increase of 33.0 (24.3) BLEU for en$\to$zh (zh$\to$en) on the full dataset. (\autoref{sec:MT performance}).
\end{itemize}

\section{Standard Approaches to Parallel Corpus Construction}
\label{sec:preliminaries}
Construction of a sentence-aligned parallel corpora from multilingual websites involves the following steps. 

\begin{enumerate}
\setlength\itemsep{0em}
\item Documents in desired languages are crawled from multi-lingual websites. 
\item Plain texts are extracted from crawled documents and normalized. 
\item Documents from both languages are matched according to their contents. 
\item Within each document, paragraphs are broken down into individual sentences.
\item Sentences are subsequently aligned into sentence pairs. 
\item Aligned sentence pairs are filtered remove duplicated and low-quality pairs.
\end{enumerate}

While the first two steps are well-established engineering tasks, the last four are under active research. For step 3, WMT16 hosted a shared task for bilingual document alignment \citep{buck2016findings}, in which the best entry relied on matching distinct biligual phrase pairs \citep{gomes2016first}. For step 4, \citet{read2012sentence} systematically evaluated nine existing tools for sentence boundary detection, among which \verb|LingPipe| \citep{lingpipe} and \verb|Punkt| \citep{loper2002nltk} are the top performers in the biomedical domain. Sentence alignment (step 5) is perhaps the most challenging among all. Compared with document alignment, sentence alignment use a smaller amount of text but more permutations. Various methods has been proposed, among which are length-based algorithm \citep{gale1993program}, lexicon-based algorithm \citep{moore2002fast,varga2007parallel,ma2006champollion}, and translation-based algorithm \citep{sennrich2011iterative,sennrich2010mt}. For step 6, WMT18 hosted a shared task on parallel corpora filtering \citep{koehn2018findings}.

\begin{figure}[t!]
\includegraphics[width=\columnwidth]{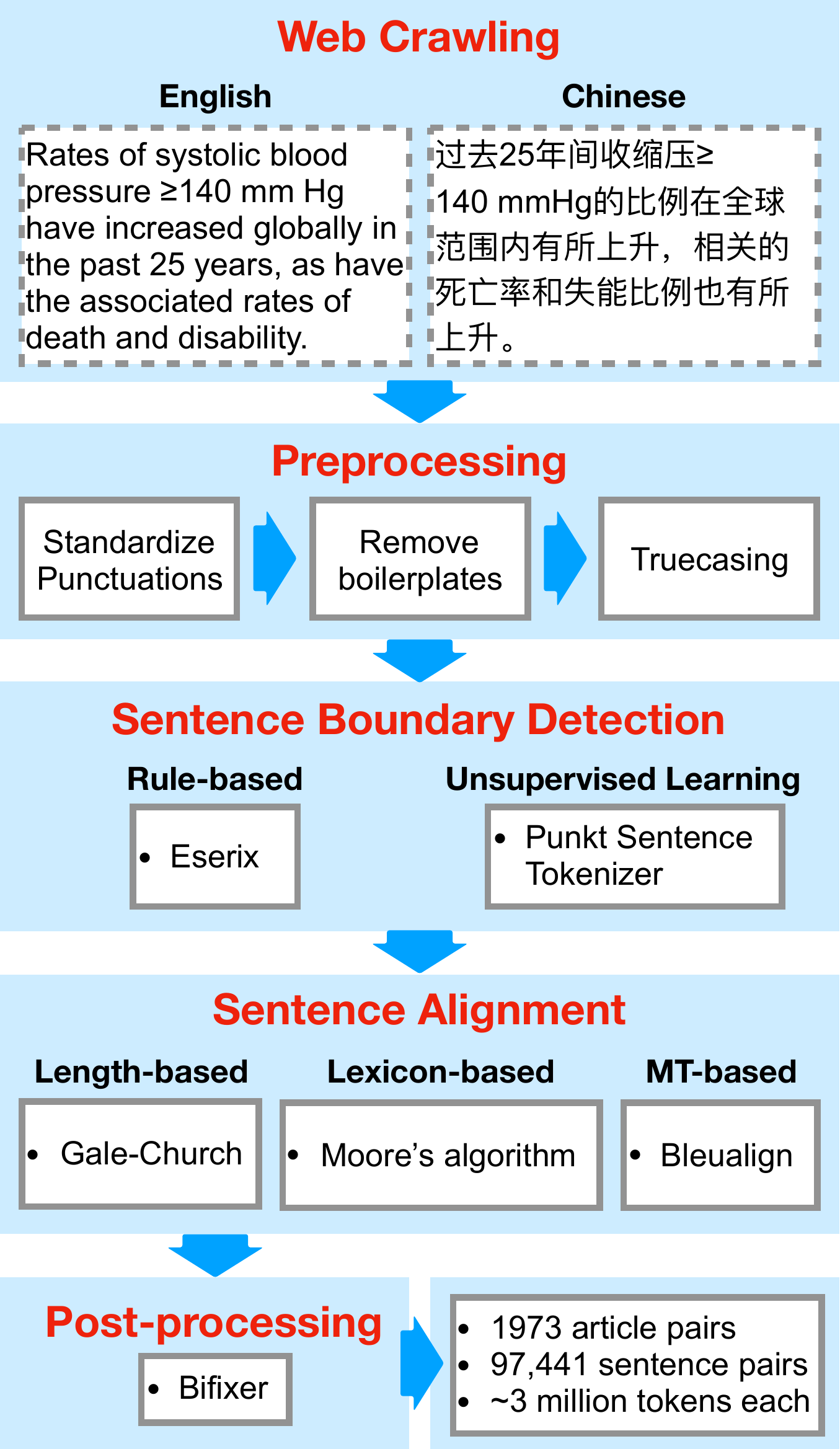}
\caption{The overall pipeline to construct the NEJM-enzh dataset. NEJM webpages were crawled using Selenium (\autoref{sec:web crawling}). Various preprocessing steps were carried out to standardize punctuations and remove boilerplate texts (\autoref{sec:preprocessing}). We tested two methods for splitting paragraphs into sentences (\autoref{sec:sbd}), and three methods to align English and Chinese sentence into translated pairs (\autoref{sec:sentence alignment}). Duplicated sentence pairs were removed at the end (\autoref{sec:postprocessing})}
\label{fig:2}
\end{figure}

The NEJM website provides hyperlinks between Chinese and English article pairs, allowing us to skip document alignment (step 3). Otherwise, we follow the best practices outlined therein and adapt them to our project (\autoref{fig:2}).

\section{Data Availability}
\label{sec:data availability}
We codebase and data are open-source via \url{https://github.com/boxiangliu/med_translation}.

\section{Corpus construction}
\label{sec:corpus construction}
This section describes the steps we took to construct the sentence-aligned corpora. The overall pipeline is illustrated in \autoref{fig:2}. 

\subsection{Web Crawling}
\label{sec:web crawling}
The Chinese outlet of New England Journal of Medicine (\url{https://www.nejmqianyan.cn/}) provides open-access Chinese translations dating back to 2011. All articles were translated sentence for sentence by professional translators, with occasional sentence concatenation and division for fluency, viz.~one English sentences split into two or more Chinese sentences and vice versa.  Translations were proofread by members of the editorial team and research articles were additionally proofread by statisticians. The Chinese translations are organized chronologically, making the content easy to crawl. Correspondent article pairs are connected via hyperlinks, eliminating the need for document alignment. 

We used Selenium \citep{salunke2014selenium} to crawl all available Chinese and corresponding English articles. While paragraph orderings are maintained across languages, locations of display items, vis. figures, tables and associated captions, are shuffled. We removed display items to keep content orders identical across English and Chinese. The English NEJM website contains untranslated auxillary contents such as job boards and visual ads. We instructed Selenium to ignore auxillary contents as these interjections makes sentence alignment challenging. Chinese NEJM translations are cleaner but contains boilerplate sentences such as names of translators. These boilerplate contents were removed during preprocessing (\autoref{sec:preprocessing}). 

\subsubsection{Article Statistics}
\begin{figure}[h]
\includegraphics[width=\columnwidth]{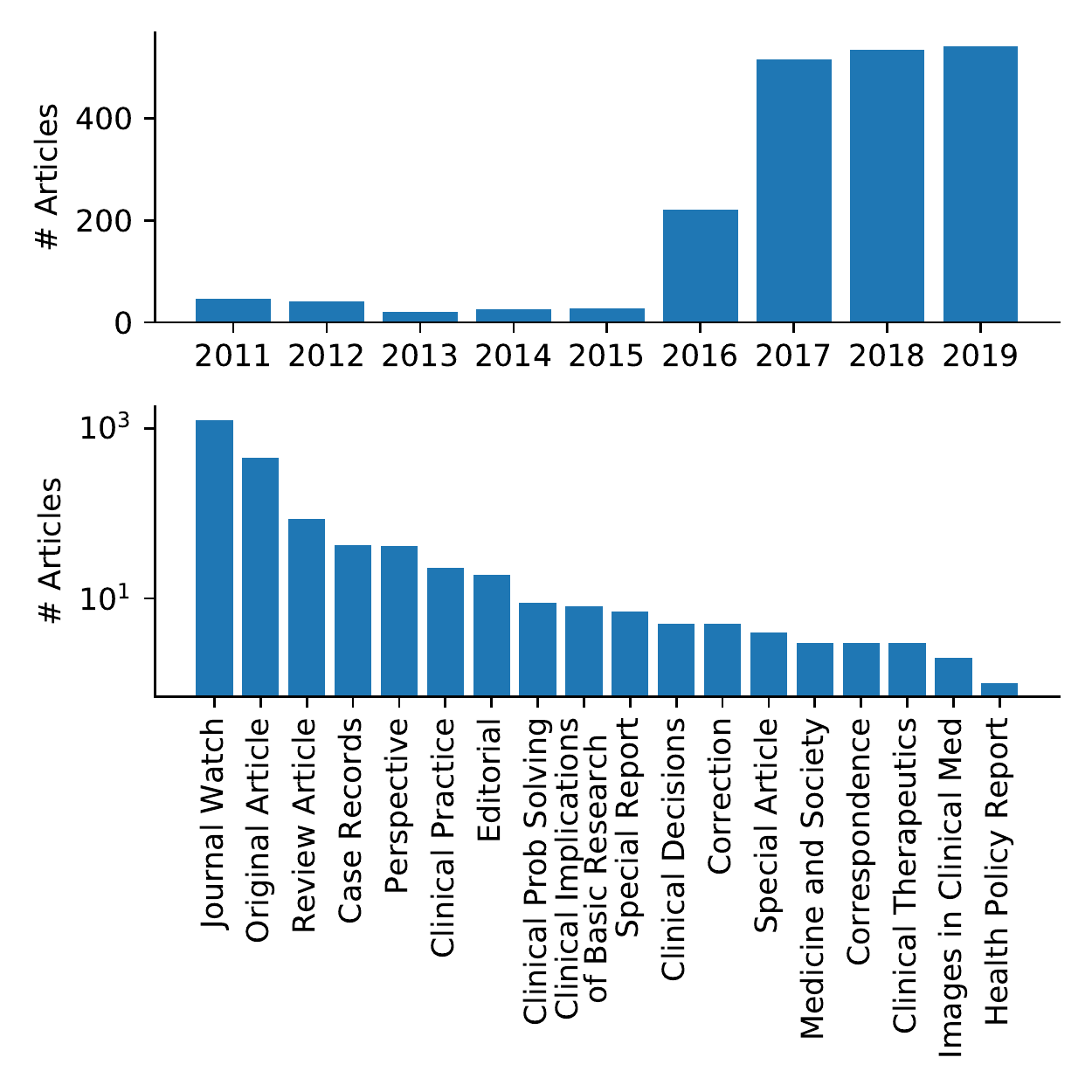}
\caption{Distribution of article by year and by type.}
\label{fig:url_statistics}
\end{figure}

NEJM has been translating more than 500 articles each year since 2016. Journal Watch (article highlight) leads in the number of articles, followed by original research and review articles. \autoref{fig:url_statistics} shows the distribution of articles by year and type.

\subsection{Preprocessing}
\label{sec:preprocessing}
We truecased letters and standardized punctuations for crawled articles with \verb|moses| \citep{koehn2007moses}. We subsequently performed stitching and filtering described below.

\subsubsection{Stitching}
A full sentence can be split incorrectly during content uploading and crawling. In Chinese articles, we found that sentence breaks can be inserted by mistake before citations and before punctuations. We assigned any text segment consisted only of citations and/or punctuations to its preceding sentence. For English, we noticed that the hyperlink phrase - open in new tab - unanimously break a full sentence into two halves. We concatenate flanking sentences and remove the hyperlink phrase. 

\subsubsection{Filtering}
We filter out the following for both languages:
\begin{itemize}[noitemsep,topsep=5pt]
	\item Figures and figure captions
	\item Tables and table legends
	\item Reference section
\end{itemize}

For Chinese, we further remove any information about translators. For English, we further remove: 

\begin{itemize}[noitemsep,topsep=5pt]
	\item Video
	\item Interactive graphic
	\item Audio interview
	\item Visual abstract
	\item Quick take (video summary)
\end{itemize}

Our codebase details all filtering steps (\autoref{sec:data availability}). 

\subsubsection{Comparison between pre- and post-filter corpora}
\begin{figure}[h]
\includegraphics[width=\columnwidth]{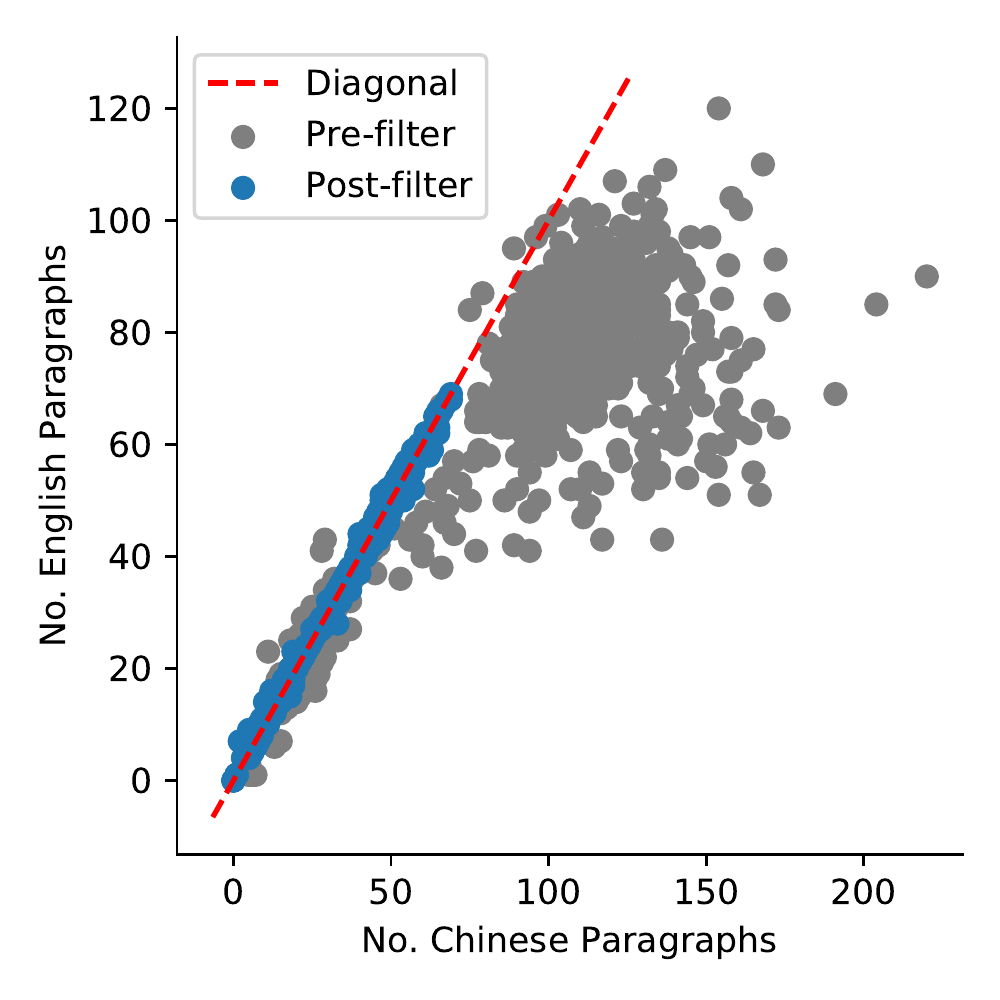}
\caption{Number of Chinese and English paragraphs are closer post-filtering compared to pre-filtering.}
\label{fig:length_comparison}
\end{figure}

\autoref{fig:length_comparison} compares the number of Chinese and English paragraphs in each article before and after filtering. Prior to filtering, the number of Chinese paragraph exceeds that of English for numerous articles, indicated by the grey sub-diagonal cloud. This is due to the various boilerplate texts in the Chinese and English articles. The number of English and Chinese paragraphs in each article become closer after filtering.

\subsection{Sentence Boundary Detection (SBD)}
\label{sec:sbd}
\begin{figure}[h]
\includegraphics[width=\columnwidth]{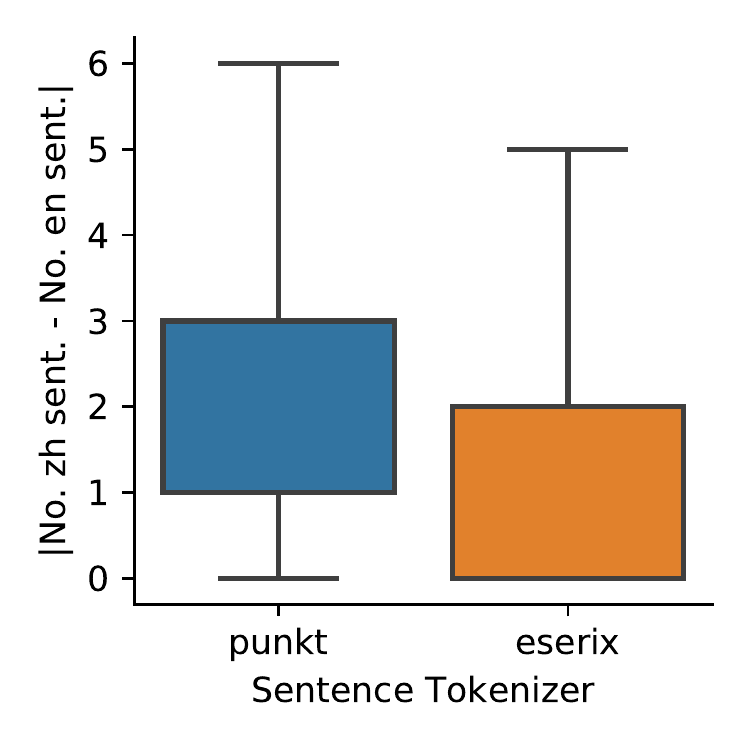}
\caption{The difference in the number of Chinese and English sentences are smallers in eserix than in punkt. Note that the distributions are left skewed and the medians overlap with the 1st quartiles.}
\label{fig:cmp_punkt_eserix}
\end{figure}

Chinese sentences are concluded by three full-stop punctuations \{！,？,。\} used exclusively for sentence separation. Unlike European languages, they do not double as decimal points or other linguistic markers. Further, Chinese quotation marks appear before sentence breaks, making it easy to detect sentence boundaries. Breaking English sentences is more challenging due to punctuation overloading. \citet{read2012sentence} showed that \verb|Punkt|, an unsupervised sentence tokenizer, performs well on biomedical corpora. We trained \verb|Punkt| on our NEJM-enzh corpus and used the learned parameters to break sentences. We also tested a rule-based system \verb|eserix| introduced by \citet{ziemski2016united} with additional rules specifically for the NEJM-enzh corpus. To compare the two, we plotted the difference in the number of sentences. Because NEJM articles were translated sentence for sentence, the ideal SBD result should have a difference of zero. We found that difference is smaller in \verb|eserix| and thus used it for downstream analysis (\autoref{fig:cmp_punkt_eserix}).

\begin{figure*}[h!]
\resizebox{\textwidth}{!}{
\setlength{\tabcolsep}{1.5pt}
\begin{tabu}{l | l l l l l l l l l l l l l }
  \rowfont{\small\itshape}
& y\v{v} & \=anw\`eij\`i & z\v{u}  & xi\=angb\v{i} & f\'uxi\`e & z\`ai & p\`atu\=ozh\=ud\=ank\`ang & z\v{u} & ji\`aow\'ei & ch\'angji\`an & \,\,\,\,\,y\'ou & hu\`of\=um\`an & lu\'osh\`i \\
(a) Chinese & 与 & 安慰剂 & 组 & 相比，& 腹泻 & 在 & 帕妥珠单抗 & 组 & 较为 & 常见 & \color{blue}（由 & \color{blue}霍夫曼-- & \color{blue}罗氏...）。\\[-0.3em]
\rowfont{\small\bfseries}
 & with & placebo & group & compared, & diarrhea & in & pertuzumab & group & relatively & common & \,\,\,\,(by & F. Hoffman & La Roche...)。\\
\hline
(b) English & \multicolumn{13}{l}{Diarrhea was more common with pertuzumab than with placebo. \color{blue}(Funded by F. Hoffmann–La Roche...).} \\
 \end{tabu}
}
\caption{Example of an erroneous break before the red text. Notice the additional period before the open parenthesis for the English text.}
\label{tab:erroneous_break}
\end{figure*}

\subsubsection{Error Analysis}
The two most frequent error made by \verb|punkt| were failure to break at citations (\autoref{tab:failure_to_break}) and erroneous breaks before open parenthesis (\autoref{tab:erroneous_break}). The latter created difficulty for sentence alignment because the Chinese sentence break appear after the close parenthesis. Conversely, \verb|eserix| did not make these mistakes. 

\begin{table}[h]
\centering
\small
\begin{tabular}{l}
No replicated loci with genomewide significance have been\\
reported\textcolor{red}{.$^{12-14}$} To overcome sample-size limitations...
\end{tabular}
\caption{Example of a failure to break two sentences due to a citation (\textcolor{red}{red text}).}
\label{tab:failure_to_break}
\end{table}

\subsection{Sentence Alignment}
\label{sec:sentence alignment}
While a number of methods has been proposed for sentence alignment, there lacks a consensus on their performance in the biomedical domain. We compared three methods from each of the three categories: length-based (Gale-Church), lexicon-based (Microsoft Aligner), and translation-based (Bleualign). The Gale-Church algorithm finds sentence pairs by assuming that the lengths of source and target sentences should be similar \citep{gale1993program}. The Microsoft Aligner integrate word correspondence with sentence length to search for sentence pairs \cite{moore2002fast}. Bleualign compareds original and translated texts to search for anchor sentences and subsequently align the rest with the Gale-Church algorithm \cite{sennrich2010mt}. To compare these methods, we establish a test set by manually aligning 1,019 sentences from 12 articles. \autoref{tab:alignment_distribution} shows the distribution of alignment types. Nearly 95\% of all alignments are one-to-one. An example of one-to-many alignment is shown in \autoref{tab:one_to_many}.

\begin{table}[ht]
\centering
\begin{tabular}{crr}
\toprule
 \textbf{zh - en} & \textbf{Count} & \textbf{Percent} \\
\midrule
 0 - 1 &     10 &   1.0\% \\
 1 - 0 &     11 &   1.1\% \\
 1 - 1 &    964 &  94.6\% \\
 1 - 2 &     17 &   1.7\% \\
 2 - 1 &     15 &   1.5\% \\
 2 - 2 &      1 &   0.1\% \\
 2 - 3 &      1 &   0.1\% \\
\bottomrule
\end{tabular}
\caption{Alignment counts in manually aligned sentence pairs, where the majority are 1-1 alignments.}
\label{tab:alignment_distribution}
\end{table}

\begin{figure*}[h!]
\resizebox{\textwidth}{!}{
\setlength{\tabcolsep}{1.5pt}
\begin{tabu}{l | l l l l l l l l l l l l l l l }
\rowfont{\small\itshape}
&  & sh\`i & y\`izh\v{o}ng & k\v{o}uf\'u & n\`a & p\'utaot\'ang & xi\'et\'ongzhu\v{a}ny\`und\`anb\'ai & y\={\i} & h\'e & \`er & de & y\`izh\`ij\`i & w\v{o}m\'en \\
(a) Chinese & \color{purple}Sotagliflozin & \color{red}是 & \color{red}一种 &\color{red}口服 & \color{red}钠-- & \color{red}葡萄糖 & \color{red}协同转运蛋白--& \color{red}1 & \color{red}和 &  \color{red}2 & \color{red}的 & \color{red}抑制剂。 &  \color{blue}我们 \\[-0.3em]
\rowfont{\small\bfseries}
& Sotagliflozin & is & an & oral & sodium- & glucose & 
cotransporters- & 1 & and & 2 & 's & inhibitor & we \\
\rowfont{\small\itshape}
& p\'ingji\`ale & z\`ai & y\={\i}x\'ing & t\'angni\`aob\`ing & hu\`anzh\v{e} & zh\=ong & li\'any\`ong & y\'id\v{a}os\`u & h\'e &  & de & \=anqu\'anx\`ing & h\'e & li\'aoxi\`ao \\
& \color{blue}评价了 &  \color{blue}在 & \color{blue}1型 & \color{blue}糖尿病 & \color{blue}患者 & \color{blue}中 & \color{blue}联用 & \color{blue}胰岛素 & \color{blue}和 & \color{purple}sotagliflozin & \color{blue}的 & \color{blue}安全性 & \color{blue}和 & \color{blue}疗效。\\[-0.3em]
\rowfont{\small\bfseries}
& evaluated & in... & type-1 & diabetes & patients & ...in & combination & insulin & and & sotagliflozin & 's & safety & and & efficacy. \\
\hline
(b) English & \multicolumn{14}{l}{\textcolor{blue}{We evaluated the safety and efficacy of} \textcolor{purple}{sotagliflozin}, \textcolor{red}{an oral inhibitor of sodium–glucose cotransporters 1 and 2}, \textcolor{blue}{in combination}} \\
 & \multicolumn{14}{l}{\textcolor{blue}{with insulin treatment in patients with type 1 diabetes.}} \\
 \end{tabu}
}
\caption{An examplar 1-to-2 alignment for clause breaking. The \textcolor{red}{red text} denotes the English clause corresponding to the first Chinese sentence. \textcolor{purple}{Sotagliflozin} is cited once in the English sentence, but repeated in two Chinese sentences.}
\label{tab:one_to_many}
\end{figure*}

\begin{figure}[ht]
\includegraphics[width=\columnwidth]{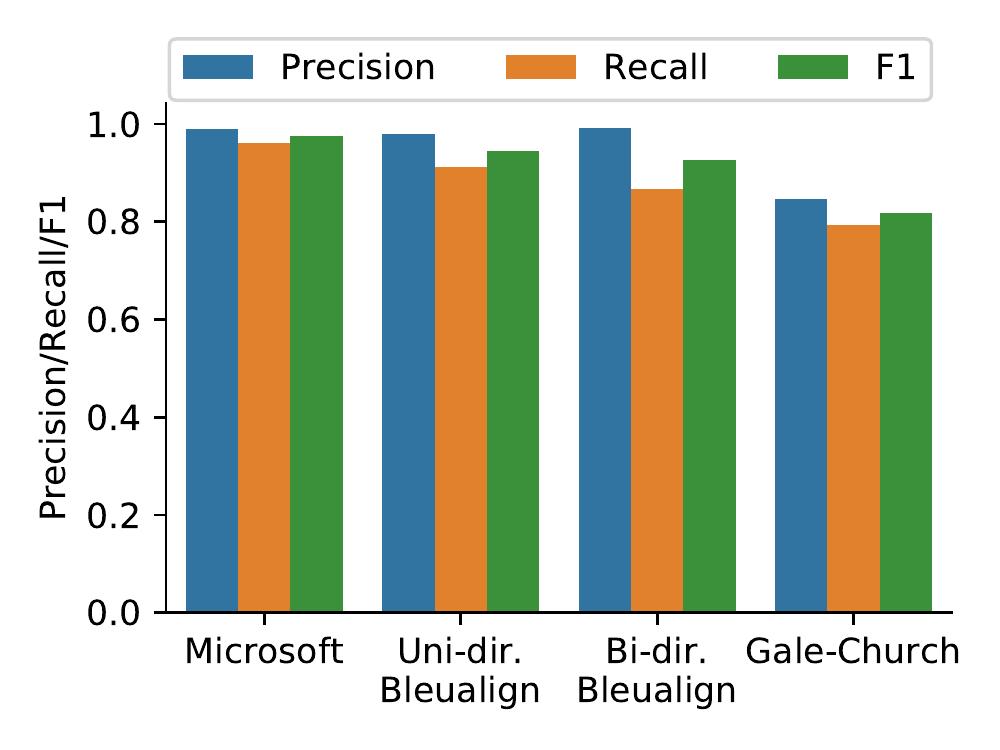}
\caption{Performance of three sentence aligners on the NEJM-enzh corpus. Uni-directional Bleualign uses translations in zh$\to$en only, whereas bi-directional Bleualign uses translations in both directions, giving it higher precision but lower recall.}
\label{fig:aligner_comparison}
\end{figure}

Because the majority of sentence pairs are one-to-one aligned, and that the performance of all algorithms degrade significantly for one-to-many alignments, we focused our attention on one-to-one alignment for this study. The precision, recall and F1 scores are shown in \autoref{fig:aligner_comparison}. The Microsoft Aligner achieved the best F1 score and was used for downstream analysis.

\subsection{Post-processing}
\label{sec:postprocessing}
Medical literature are highly structured. Certain sections including the abstract, introduction, methods, results and discussion are almost universal across articles. We remove duplicated header and other repeated text with bifixer \cite{sanchez2018prompsit}.

\subsection{Training and test split}
We selected 2,102 sentences from 39 latest articles as the test set and 2,036 sentences from the next latest 40 articles as the development set. To avoid data leakage, full articles must be in either train, development and test set.

\section{Translation quality}
\label{sec:MT performance}
To measure the effect that the NEJM-enzh corpus has on medical translation, we compared a baseline transformer model trained on the WMT18 English-Chinese dataset and one fine-tuned with the NEJM-enzh corpus. Although translations are evaluated bidirectionally, it should be emphasized that the NEJM-enzh corpus is translated from English to Chinese and this will influence the machine translation quality \citep{graham2019translationese}. 

\subsection{Model Architecture}
We used the transformer model in OpenNMT with 6 layers, each with an output size of 512 hidden units \cite{klein-etal-2017-opennmt}. We used 8 attention heads and sinusoidal positional embedding. The final hidden feed-forward layer is of size 2,048.

\subsection{Training Data}
We trained a baseline model on the English-Chinese parallel corpus from WMT18 \cite{bojar-EtAl:2018:WMT1} consisting about 24.8 million sentence pairs. Sentences are encoded with Byte-Pair Encoding \cite{sennrich2015neural} with vocabularies of 16,000 tokens for each language. Sentence lengths are capped at 999 tokens, enough to accommodate most sentences.

\subsection{Hardware and training procedure}
We trained our model on 8 Nvidia TitanX GPUs. We used the Adam optimizer \cite{kingma2014adam} with $\beta_1 = 0.9$ and $\beta_2 = 0.997$ and 10,000 warm-up steps. We applied dropout with $p_{d} = 0.1$ and label smoothing with $\epsilon_{ls} = 0.1$. The model was trained for 500,000 steps in total. The training procedure took 4.5 days. We fine-tuned the baseline model on NEJM-enzh for 100,000 steps with identical parameters. To establish a second comparison, we trained a transformer model \textit{de novo} on the NEJM-enzh corpus.

\subsection{Results}
\begin{figure}[ht]
\centering
\includegraphics[width=\columnwidth]{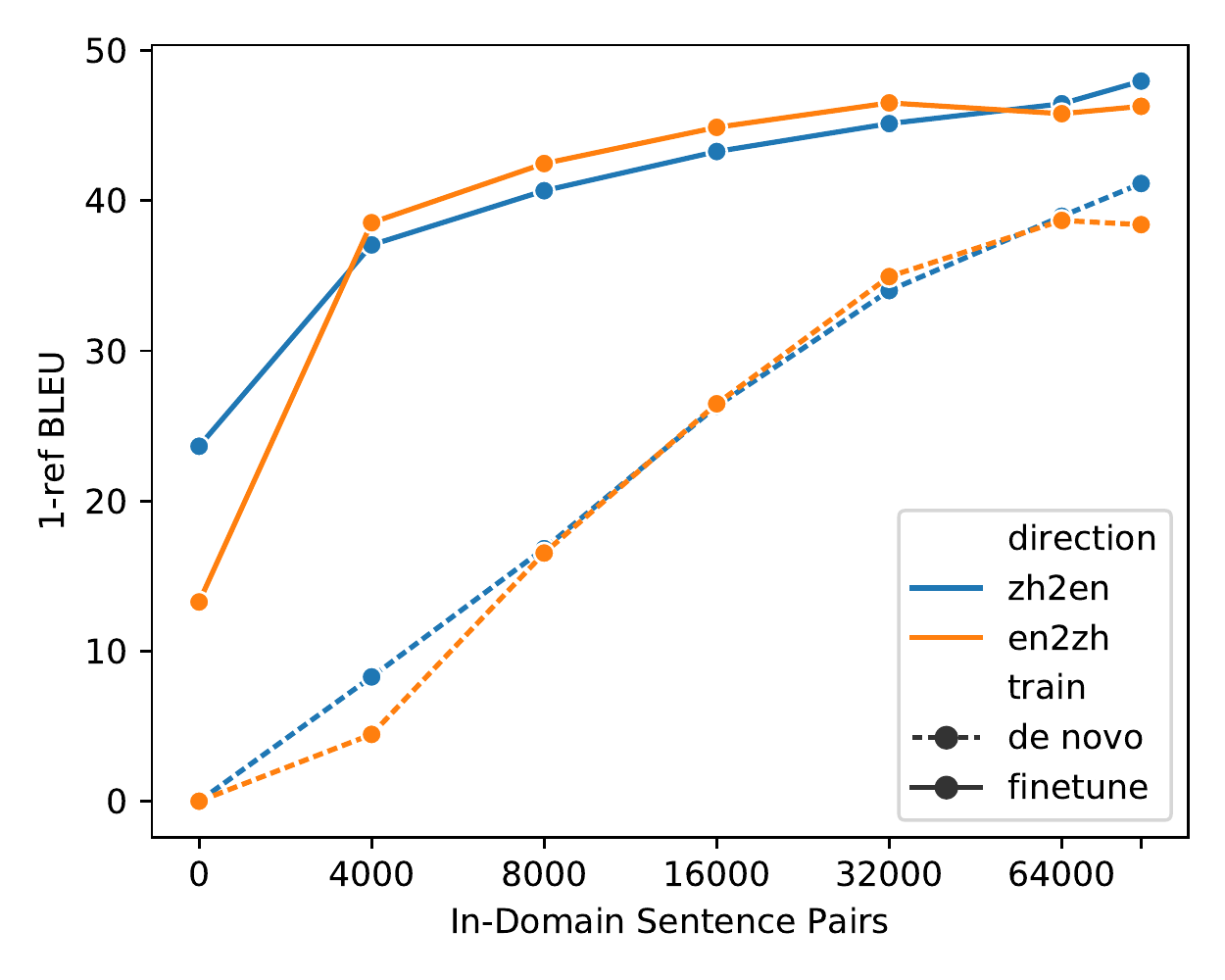}
\caption{Increase in translation quality by increasing in-domain corpus size.}
\label{fig:bleu_score}
\end{figure}

To understand the translation quality as a function of in-domain dataset size, we fine-tuned our model on 4,000, 8,000, 16,000, 32,000, 64,000 and all 93303 sentence pairs (\autoref{fig:bleu_score}). For both zh$\to$en and en$\to$zh models, we saw improvement as the number of in-domain sentence pairs increased. The most significant improvement occured at 4,000 sentence pairs (en$\to$zh: +25.3 BLEU; zh$\to$en: +13.4 BLEU). Translation quality continued to improve as the size of dataset grows, albeit at a slower pace. Compared with baseline, the full dataset with 93303 sentence pairs increased the BLEU score by 33.0 (24.3) points in en$\to$zh (zh$\to$en) directions.

To determine whether the pre-training on WMT18 newswire data is necessary, we trained a \textit{de novo} model using only NEJM-enzh data, which was significantly faster than training a baseline model followed by fine-tuning. Compared with \textit{de novo} training, pre-training on WMT18 baseline plus fine-tuning provided a meaningful boost in translation quality. Such a boost was most evident on small in-domain dataset. With 4,000 sentence pairs, pre-training improved the BLEU score by 34.1 (28.8) points for en$\to$zh (zh$\to$en) directions. The difference remained but decreased as in-domain dataset grew, dropping to 7.9 (6.8) BLEU for en$\to$zh (zh$\to$en) at the full-set level. A larger in-domain dataset is needed to completely compensate the translation quality drop without pre-training.

\subsection{Error Analysis}
We show two examples in this section to illustrate common mistakes made by our models. In the zh$\to$en direction, the phrase ``铂类-紫杉类" was correctly translated by the fine-tuned model to ``platinum-taxane", and mistranslated by the baseline model to ``Pt-Pseudophyllus" （\autoref{tab:error_analysis_1}). The baseline model has not seen the phrase ``紫杉类" during training and thus resulted in incorrect decoding. A similar situation occured at the phrase ``贝伐珠单抗". The fine-tuned model was able to correctly translate the phrase into bevacizumab, a chemotherapy medication, whereas the baseline model incorrectly decoded the phrase as ``Bavaris mono-repellent". 

Similar situations occured for the en$\to$zh direction (\autoref{tab:error_analysis_2}). Two medications, ``olaparib" and ``bevacizumab", were correctly translated by the fine-tuned model as ``奥拉帕利" and ``贝伐珠单抗", but incorrectly translated by the baseline model as ``孤寡老人" and ``白蜂"。

Fine-tuning on in-domain data extended the model vocabulary and made it more accurate to decode medical terminology.

\begin{figure*}[h!]
\resizebox{\textwidth}{!}{
\setlength{\tabcolsep}{1.5pt}
\begin{tabu}{l | l l l l l l l l l l l }
\rowfont{\small\itshape}
& hu\`anzh\v{e} & ji\=esh\`ou & b\'ol\`ei & z\v{\i}sh\=anl\`ei & y\`aow\`u & hu\`ali\'ao & b\`eif\'azh\=ud\=ank\`ang & y\={\i}xi\`an & zh\`ili\'ao & h\`ou\\
(a) Source & 患者 & 接受 & \color{blue}铂类--& \color{blue}紫杉类 & 药物 & 化疗+ & \color{blue}贝伐珠单抗 & 一线 & 治疗 & 后, & \\[-0.3em]
\rowfont{\small\bfseries}
& patients & receiving & platinum & taxane & drug & chemotherapy & bevacizumab & first-line & treatment & afterwards \\

\rowfont{\small\itshape}
& b\v{e}n & y\'anji\=u & y\=aoqi\'u & q\'i & b\`un\'eng & y\v{o}u & b\`ingbi\`an & j\`ixi\`ang & hu\`ozh\v{e} & z\`ai & zh\`ili\'ao \\
& 本 & 研究 & 要求 & 其 & 不能 & 有 & 病变 & 迹象, & 或者 & 在 & 治疗 \\[-0.3em]
\rowfont{\small\bfseries}
& this & study & require & they & don't & have & disease & evidence & or & after... & treatment \\

\rowfont{\small\itshape}
& h\`ou  & d\'ad\`ao & l\'inchu\'ang & w\'anqu\'an & hu\`o & b\`ufen & hu\v{a}nji\v{e} & \,\,\,\,\,d\`ingy\`i & c\=anji\`an & bi\v{a}o \\
& 后 & 达到 & 临床 & 完全 & 或 & 部分 & 缓解 & （定义 & 参见 & 表1）。& \\[-0.3em]
\rowfont{\small\bfseries}
& ...after & achieve & clinical & complete & or & partial & relief & definition & see & table 1\\

\hline
(b) Target & \multicolumn{11}{l}{After first-line treatment with \textcolor{blue}{platinum-taxane} chemotherapy plus \textcolor{blue}{bevacizumab}, patients were required to} \\
& \multicolumn{11}{l}{have no evidence of disease or to have had a clinical complete or partial response (definitions in Table 1).} \\

\hline
(c) NEJM & \multicolumn{11}{l}{Patients were required to have no evidence of disease or to have a clinical complete or partial response} \\ 
translation & \multicolumn{11}{l}{after treatment after first-line \textcolor{blue}{platinum-taxane} chemotherapy plus \textcolor{blue}{bevacizumab} (as defined in Table 1).} \\
\hline
(d) WMT18 & \multicolumn{11}{l}{after \textcolor{red}{Pt-Pseudophyllus} drug chemotherapy + \textcolor{red}{Bavaris mono-repellent}  first-line treatment, the study required that the} \\
translation & \multicolumn{11}{l}{patient should not show signs of lesion or complete or partial clinical relief after treatment (see table 1 for definition).} \\

 \end{tabu}
}
\caption{\textcolor{blue}{铂类-紫杉类} and \textcolor{blue}{贝伐珠单抗} were never seen by the baseline model and were translated incorrectly (\textcolor{red}{red text}).}
\label{tab:error_analysis_1}
\end{figure*}

\begin{figure*}[h!]
\resizebox{\textwidth}{!}{
\setlength{\tabcolsep}{1.5pt}
\begin{tabu}{l | l l l l l l l l l l l l l l}
(a) Source & \multicolumn{14}{l}{The lack of a maintenance \textcolor{blue}{olaparib} monotherapy comparator group is a limitation of the PAOLA-1 trial, making it difficult to conclude whether the progression-free}\\
& \multicolumn{14}{l}{survival benefit seen in patients with HRD-positive tumors without BRCA mutations (who were not included in the SOLO1 trial) was due largely to the \textcolor{blue}{addition of}}\\
& \multicolumn{14}{l}{\textcolor{blue}{olaparib} or whether a synergistic effect occurred with \textcolor{blue}{olaparib} and \textcolor{blue}{bevacizumab}.}\\

\hline

\rowfont{\small\itshape}
& w\`eish\`ezh\`i & \`aol\=ap\`al\`i & d\=any\`ao & w\'eich\'i & zh\`ili\'ao & d\`uizh\`aoz\v{u} & sh\`i &  & sh\`iy\`an & de & y\'ig\`e & j\'uxi\`anx\`ing & zh\'e & sh\v{\i}d\'e \\
(b) Target & 未设置 & \color{blue}奥拉帕利 & 单药 & 维持 & 治疗 & 对照组 & 是 & PAOLA-1 & 试验 & 的 & 一个 & 局限性, & 这 & 使得 \\[-0.3em]
\rowfont{\small\bfseries}
& lacking & olaparib & monotherapy & maintenance & treatment & comparator & is & PAOLA-1 & trial & 's & a & limitation & this & makes \\
 
\rowfont{\small\itshape}
 & w\v{o}m\=en & n\'any\v{\i} & qu\`ed\`ing & z\`ai & w\'u &  & t\=ubi\`an & de & & y\'angx\`ing & zh\v{o}ngli\'u & hu\`anzh\v{e} & & sh\`iy\`an \\
 & 我们 & 难以 & 确定 & 在 & 无 & BRCA & 突变 & 的 & HRD & 阳性 & 肿瘤 & 患者 & (SOLO1 & 试验 \\[-0.3em]
 \rowfont{\small\bfseries}
 & us & difficult to & determine & in... & no & BRCA & mutation & 's & HRD & positive & tumor & patients & (SOLO1 & trial \\
 
 \rowfont{\small\itshape}
 & w\`ei & n\`ar\`u & c\v{\i}l\`ei & hu\`anzh\v{e} & zh\=ong & gu\=anch\'ad\`ao & w\'uj\`inzh\v{a}n & sh\=engc\'unq\={\i} & hu\`oy\`i & de & yu\'any\={\i}n & sh\`if\v{o}u & zh\v{u}y\`aosh\`i \\
 & 未 & 纳入 & 此类 & 患者) & 中 & 观察到 & 无进展 & 生存期 & 获益 & 的 & 原因 & 是否 & 主要是 \\[-0.3em]
 \rowfont{\small\bfseries}
 & not & including & these & patients) & ...in & observed & prog.-free & survival & benefit & 's & reason & whether & largely \\
 
 \rowfont{\small\itshape}
 & y\'ouy\'u & ji\=ay\`ong & \`aol\=ap\`al\`i & y\v{e} & w\'uf\v{a} & qu\`ed\`ing & \`aol\=ap\`al\`i & h\'e & b\`eif\'azh\=ud\=ank\`ang & sh\`if\v{o}u & ch\v{a}nsh\=eng & xi\'et\'ong & zu\`oy\`ong \\
 & 由于 & \color{blue}加用 & \color{blue}奥拉帕利, & 也 & 无法 & 确定 & \color{blue}奥拉帕利 & 和 & \color{blue}贝伐珠单抗 & 是否 & 产生 & 协同 & 作用. \\[-0.3em]
 \rowfont{\small\bfseries}
 & due to & adding & olaparib & also & can't & determine & olaparib & and & bevacizumab & whether & produce & synergistic & effect. \\

\hline
\rowfont{\small\itshape}
& qu\=ef\'a & \`aol\=ap\`al\`i & w\'eich\'i & d\=any\`ao & zh\`ili\'ao & d\`uizh\`aoz\v{u} & sh\`i & & sh\`iy\`an & de & y\'ig\`e & j\'uxi\`anx\`ing & y\={\i}nc\v{\i} & n\'any\v{\i} \\
(c) NEJM & 缺乏 & \color{blue}奥拉帕利 & 维持 & 单药 & 治疗 & 对照组 & 是 & PAOLA-1 & 试验 & 的 & 一个 & 局限性, & 因此 & 难以 \\[-0.3em]
\rowfont{\small\bfseries}
translation & lacking & olaparib & maintenance & monotherapy & treatment & comparator & is & PAOLA-1 & trial & 's & a & limitation, & therefore & difficult to \\

\rowfont{\small\itshape}
 & qu\`ed\`ing & z\`ai & w\'u &  & t\=ubi\`an & de & & y\'angx\`ing & zh\v{o}ngli\'u & hu\`anzh\v{e} & w\`ei & n\`ar\`u & & sh\`iy\`an \\
 & 确定 & 在 & 无 & BRCA & 突变 & 的 & HRD & 阳性 & 肿瘤 & 患者 & (未 & 纳入 & SOLO1 & 试验)\\[-0.3em]
\rowfont{\small\bfseries}
& determine & in & no & BRCA & mutation & 's & HRD & positive & tumor & patients & (not & included in & SOLO1 & trial) \\

\rowfont{\small\itshape}
& zh\=ong & gu\=anch\'ad\`aode & w\'uj\`inzh\v{a}n & sh\=engc\'unq\={\i} & hu\`oy\`i & zh\v{u}y\`ao & sh\`i & y\'ouy\'u & ji\=ay\`ongle & \`aol\=ap\`al\`i & h\'aish\`i & y\'ouy\'u & \`aol\=ap\`al\`i & h\'e \\
& 中 & 观察到的 & 无进展 & 生存期 & 获益 & 主要 & 是 & 由于 & \color{blue}加用了 & \color{blue}奥拉帕利, & 还是 & 由于 & \color{blue}奥拉帕利 & 和 \\[-0.3em]
\rowfont{\small\bfseries}
& in & observed & prog.-free & survival & benefit & largely & is & due to & adding & olaparib, & or & due to & olaparib & and \\

\rowfont{\small\itshape}
& b\`eif\'azh\=ud\=ank\`ang & ch\v{a}nsh\=engle & xi\'et\'ong & zu\`oy\`ong \\
 & \color{blue}贝伐珠单抗 & 产生了 & 协同 & 作用. \\[-0.3em]
\rowfont{\small\bfseries}
 & bevacizumab & produced & synergistic & effect. \\
\hline

\rowfont{\small\itshape}
& qu\=ef\'a & w\'eich\'i & g\=ugu\v{a} & l\v{a}or\'en & d\=any\={\i} & li\'aof\v{a} & b\v{\i}ji\`ao & xi\v{a}oz\v{u} & sh\`i & & sh\`iy\`an & de & y\'ig\`e & xi\'anzh\'i \\
(d) WMT18 & 缺乏 & 维持 & \color{red}孤寡 & \color{red}老人 & 单一 & 疗法 & 比较 & 小组 & 是 & PAOLA-1 & 试验 & 的 & 一个 & 限制, \\[-0.3em]
\rowfont{\small\bfseries}
translation & lacking & maintenance & lonely & elderly & mono & therapy & 
comparator & group & is & PAOLA-1 & trial & 's & a & limitation, \\
\rowfont{\small\itshape}
& sh\v{\i}d\'e & n\'any\v{\i} & du\`and\`ing & z\`ai & m\'eiy\v{o}u & & t\=ubi\`an & de & & y\'angx\`ing & zh\v{o}ngli\'u & hu\`anzh\v{e} & w\`ei & b\=aoku\`o \\
 & 使得 & 难以 & 断定 & 在 & 没有 & BRCA & 突变 & 的 & HRD & 阳性 & 肿瘤 & 患者 & (未 & 包括 \\[-0.3em]
 \rowfont{\small\bfseries}
 & making & difficult to & determine & in... & no & BRCA & mutation & 's & HRD & positive & tumor & patients & (not & included \\

\rowfont{\small\itshape}
& z\`ai & & zh\`iy\`anzh\=ong & zh\=ong & k\`and\`aode & w\'uj\`inzh\v{a}n & sh\=engc\'un & l\`iy\`i & zh\v{u}y\'ao & sh\`i & y\'ouy\'u & gu\v{a}t\'ou & y\`ouch\'ong & de \\
& 在 & SOLO1 & 试验中）& 中 & 看到的 & 无进展 & 生存 & 利益 & 主要 & 是 & 由于 & \color{red}寡头 & \color{red}幼虫 & \color{red}的 \\[-0.3em]
\rowfont{\small\bfseries}
& in & SOLO1 & trial) & ...in & observed & prog.-free & survival & benefit & largely & is & due to & oligo- & larva & 's \\

\rowfont{\small\itshape}
& z\=engji\=a & h\'aish\`i & y\v{u} & gu\v{a}t\'ou & h\'e & b\'ai & f\=eng & f\=ash\=eng & xi\'et\'ong & zu\`oy\`ong. \\
& \color{red}增加 & 还是 & 与 & \color{red}寡头 & 和 & \color{red}白 & \color{red}蜂 & 发生 & 协同 & 作用.\\[-0.3em]
\rowfont{\small\bfseries}
& increase\color{blue} & or & with & oligo- & and & white & bee & produce & synergistic & effect. \\

 \end{tabu}
}
\caption{\textcolor{blue}{Olaparib} and \textcolor{blue}{bevacizumab} were not seen by the baseline model and were translated incorrectly (\textcolor{red}{red text}).}
\label{tab:error_analysis_2}
\end{figure*}

\section{Related Work}
\label{sec:related work}
\subsection{Parallel texts in biomedical domain}
A number of parallel corpora are available in the public domain. The UFAL Medical Corpus covers language pairs from English to Czech, German, Spanish, French, Hungarian, Polish, Romainian and Swedish. The corpus is a collection of smaller datasets such as Medical Web Crawl and CzEng. 

The ReBEC dataset \citep{neves2017parallel} contains Portugeuse and English parallel text obtained from 1,188 clinical trial documents in Brazilian Clinical Trials Registry.

The WMT19 Biomedical Translation Workshop organizers \citep{bawden2019findings} provide sentence pairs from Medline abstracts between English and Spanish, German, Portugese and French, as well as from the EDP database between English and French.

The Khresmoi dataset \citep{11234/1-2122} samples 1,500 English sentences from medical documents. These sentences are manually translated into Czech, French, German, Hungarian, Polish, Spanish and Swedish. This dataset is ideally used as a development dataset because of its small size, high translation quality, and wide coverage of languages.

The MeSpEn dataset \citep{villegas2018mespen} contains English and Spanish parallel text collected from IBECS (Spanish Bibliographical Index in Health Sciences), SciELO (Scientific Electronic Library Online), Pubmed and MedlinePlus. In addition, it also collected 46 bilingual medical glossaries for a number of language pairs besides English and Spanish. 

The Unified Medical Language System contains medical lexicon across numerous languages but does not offer parallel sentences \citep{bodenreider2004unified}. 

\subsection{Machine Translation in the Biomedical Domain} 
The popularity of neural machine translation models has boosted the need for large datasets. Public releases of many large-scale parallel corpora have significantly improved the quality of machine translation. As an example, coordinated efforts such as the OPUS project have released multiple datasets each year across many domains and languages \citep{TIEDEMANN12.463}.

Machine translation in the biomedical domain has seen increasing attention in recent years \citep{bojar2016findings,barrault2019findings,Bojar2017findings}. Apart from other domains, biomedical literature are rich in terminology to describe various diseases and biological processes. The Unified Medical Language System (UMLS) contains over 2 millions tokens for over 900,000 medical concepts, many of which rarely appear in existing parallel corpora \citep{bodenreider2004unified}. To add to this challenge, biomedical translation mandates high standard of translation accuracy as the consequence of misinterpretation in medical decisions can be severe. All these challenges call for the curation of large-scale biomedical parallel corpora.

Despite the need for biomedical parallel text, curation of large-scale corpora has been lacking. Because of the domain knowledge needed for accurate translation of medical literature, very few have attempted this task. Biomedical parallel corpora have been made available across several pairs of European languages, including English, German, Spanish, France, Portugese, and Polish, to name a few. For Chinese, two studies have described pipelines for constructing biomedical parallel corpora, but neither have released data to the public \citep{chen2011construction,tang2018building}. Because of this, the latest shared task on WMT19 at the time of this writing did not provide in-domain training data for English-Chinese translation \citep{barrault2019findings}.

\subsection{Domain Adaptation in Machine Translation}
Domain adaptation refers to trainining machine translation models using out-of-domain (OOD) data. Two prevailing challenges hinders training with OOD data. Instances that have not appeared in the training set (covariate shift) are difficult to translate accurately during test time. Medical terminology, such as the word ``oncogenesis", that rarely occurs in the general domain falls into this category. Instances that appear during training and testing with different semantics present another challenge. For example, ``clinical \textit{depression}" is translated to Spanish as ``\textit{depresión} clínica" whereas ``economic \textit{depression}" is most commonly translated as ``\textit{recesión} económica".

Various domain adaptation techniques has been developed. Synthetic data generation such as forward and backward translation \citep{sennrich2015improving} aims to augment OOD parallel data with monolingual in-domain (ID) data. Data selection method aims to select in-domain examples from general domain data \citep{duh2013adaptation}. Fine-tuning with a small amount of in-domain data has been shown to substantially improve translation quality \citep{luong2015stanford}.

\section{Conclusion}
We have presented a English-Chinese parallel dataset in the biomedical domain. We have shown that a baseline model trained on WMT18 has limited generalizability to the biomedical domain, and that as few as 4,000 sentence pairs from the NEJM-enzh dataset substantially improved translation quality. The translation quality continued to improve as the dataset grew. In addition, pre-training with the out-of-domain data benefited translation quality, even at the full-set level.

\section*{Acknowledgments}
We thank Renjie Zheng, Baigong Zheng, Mingbo Ma, and Kenneth Church for their insights. 

\bibliography{anthology,acl2020}
\bibliographystyle{acl_natbib}

\end{CJK}
\end{document}